# A Data-Driven Machine Learning Approach for Predicting Axial Load Capacity in Steel Storage Rack Columns

Bakhtiyar MAMMADLI[1], Casim YAZICI[2,*], Muhammed GÜRBÜZ[3], İrfan KOCAMAN[3],
F. Javier DOMÍNGUEZ-GUTIÉRREZ[1], Fatih Mehmet ÖZKAL[4]

[1]NOMATEN Centre of Excellence, National Centre for Nuclear Research, ul. A. Soltana 7, 05-400 Otwock, Poland
[2]Department of Construction, Ağrı İbrahim Çeçen University, 04400 Ağrı, Türkiye
[3]Department of Civil Engineering, Erzurum Technical University, 25050 Erzurum, Türkiye
[4]Department of Civil Engineering, Atatürk University, 25240 Erzurum, Türkiye
cyazici@agri.edu.tr

**Abstract.** In this study, we present a machine learning (ML) framework to predict the axial load-bearing capacity, (kN), of cold-formed steel structural members. The methodology emphasizes robust model selection and interpretability, addressing the limitations of traditional analytical approaches in capturing the nonlinearities and geometrical complexities inherent to buckling behavior. The dataset, comprising key geometric and mechanical parameters of steel columns, was curated with appropriate preprocessing steps including removal of non-informative identifiers and imputation of missing values. A comprehensive suite of regression algorithms—ranging from linear models to kernel-based regressors and ensemble tree methods—was evaluated. Among these, Gradient Boosting Regression exhibited superior predictive performance across multiple metrics, including the coefficient of determination ($R^2$), root mean squared error (RMSE), and mean absolute error (MAE), and was consequently selected as the final model. Model interpretability was addressed using SHapley Additive exPlanations (SHAP), enabling insight into the relative importance and interaction of input features influencing the predicted axial capacity. To facilitate practical deployment, the model was integrated into an interactive, Python-based web interface via Streamlit. This tool allows end-users—such as structural engineers and designers—to input design parameters manually or through CSV upload, and to obtain real-time predictions of axial load capacity without the need for programming expertise. Applied to the context of steel storage rack columns, the framework demonstrates how data-driven tools can enhance design safety, streamline validation workflows, and inform decision-making in structural applications where buckling is a critical failure mode.

## 1. INTRODUCTION

Recent advances in machine learning (ML) have underscored its potential to enhance the prediction of structural performance and to improve safety in engineering applications [1-3]. Numerous studies have demonstrated the effectiveness of ML-based models in estimating critical mechanical properties of structural components [1-9]. For example, Gürbüz and Kazaz [8] employed ensemble learning methods—such as XGBoost and Random Forest—to develop predictive models for the ultimate drift ratio of steel plate shear walls, achieving high accuracy. Similarly, ensemble-based approaches have gained prominence for their superior predictive capabilities across various structural scenarios. Yazici and Domínguez-Gutiérrez [9] developed a framework utilizing Gradient Boosting, XGBoost, and Random Forest algorithms to estimate the high-temperature mechanical properties of high-strength steels, highlighting the enhanced reliability of ensemble models over traditional regression techniques.

Steel storage rack systems are essential elements in modern industrial and commercial logistics, supporting efficient material handling, inventory management, and space optimization. Traditionally, these systems have been designed based on static engineering principles, with emphasis on load-bearing capacity, structural safety, and spatial efficiency. However, as global trade expands and supply chains grow more complex, storage environments are increasingly subject to dynamic conditions, including variable inventory sizes, fluctuating demand, and real-time operational adjustments. These evolving challenges have exposed the limitations of conventional design and analysis approaches. In response, the integration of machine learning into storage rack design and operation is emerging as a transformative solution. ML enables systems to leverage large datasets, recognize complex patterns, and adapt to changing operational conditions. For instance, Chen and Barg [7] introduced the Rack-Aware regenerating code model to optimize fault-tolerant data recovery in distributed storage systems, reducing bandwidth requirements and enhancing system resilience. In the context of physical storage systems, ML has also been applied to improve automated storage and retrieval systems (AS/RS). Wang and Chen [10, 11] demonstrated the effectiveness of an ML-enhanced heuristic for optimizing unit-load placement and retrieval, thereby reducing congestion and minimizing delays in high-demand environments. These applications illustrate the broader potential of ML in addressing challenges related to structural efficiency and reliability. By integrating predictive analytics into the design and assessment of steel storage

racks—particularly in the context of local and global buckling—data-driven approaches can supplement or even outperform traditional methods, offering both higher accuracy and adaptability to diverse operating conditions.

The Archivist mechanism, introduced by Ren et al., further highlights ML's role in data management by employing data placement algorithms that optimize storage based on real-time system performance metrics and access patterns. This dynamic adjustment reduces retrieval times while optimizing storage costs, enabling more efficient data center operations [8, 9, 12]. In cloud-based environments, ML-driven automatic storage space recommendation systems further exemplify this trend. By analyzing incoming data characteristics, these systems can recommend optimal storage configurations tailored to specific data types, ensuring agile and responsive storage infrastructure. Mondal et al. demonstrated how such solutions enhance storage efficiency and scalability, especially in dynamic data environments [13]. Predictive maintenance also benefits significantly from ML integration. ML models can identify performance anomalies and detect early signs of equipment failure by analyzing historical performance data. Gao and Lu explored this capability in energy storage devices, where ML-based models predicted maintenance needs, reduced downtime, and extended equipment lifespan [14]. This predictive approach offers substantial operational savings, making maintenance interventions both proactive and cost-effective. Energy management has similarly benefited from ML-enabled predictive models. Reddy et al. introduced an intelligent energy management framework that optimizes electricity distribution using ML algorithms. Applying similar methods to AS/RS systems can significantly reduce operational costs while enhancing environmental sustainability by minimizing energy consumption [15].

Security and data integrity also receive robust support from ML-powered anomaly detection algorithms. These algorithms can identify potential security threats and data breaches in real time, ensuring the confidentiality and reliability of stored data. Rahmaty et al demonstrated how ML-based threat detection models efficiently safeguard storage systems by rapidly detecting and mitigating unauthorized access attempts [16]. Additionally, integrating blockchain technology with ML further enhances data integrity by supporting decentralized storage systems, as outlined by Kumar et al. [17]. Blockchain-enabled ML frameworks ensure secure data storage, efficient data retrieval, and tamper-proof record-keeping [15]. The convergence of ML and storage rack systems aligns with broader technological shifts toward autonomous and self-managing infrastructures. Reinforcement learning algorithms enable self-adjusting storage systems capable of adapting to fluctuating inventory levels and operational demands. Noel et al. explored this potential in self-managing cloud environments, demonstrating how autonomous systems improve responsiveness and adaptability in fast-changing storage scenarios [16]. Looking ahead, research efforts will likely focus on refining existing ML algorithms and exploring new applications within storage rack systems. Hybrid models combining multiple ML techniques could unlock further efficiencies, creating more intelligent, adaptive, and scalable storage infrastructures. As technology advances, ML's role in storage rack system optimization will expand, driving operational excellence and setting new industry benchmarks [17-19].

In conclusion, integrating ML into storage rack systems presents transformative opportunities for enhancing operational performance, optimizing resource utilization, and improving overall system functionality. By leveraging ML's predictive capabilities, businesses can deploy more adaptive, efficient, and scalable storage solutions. As advancements continue, the potential for ML-driven storage management will only expand, driving the evolution of more intelligent and sustainable storage systems. In this context, the present study introduces a machine learning-based framework specifically developed to predict the axial axial load capacity of structural steel rack columns. By combining regression models with explainable ML techniques and a user-friendly interface, this work bridges data-driven prediction and practical engineering application.

## 2. METHODOLOGY

This section outlines the methodological approach used to develop ML-based models for predicting the load-bearing capacity of storage rack systems. The process includes data collection and preprocessing [20], data analysis and visualization, model development, hyperparameter optimization, and performance evaluation.

### 2.1. Data Collection and Preprocessing

The dataset employed in this study comprises 261 storage rack column samples, each characterized by ten structural parameters known to influence load-bearing capacity. These features encompass a combination of geometric design variables, material properties, and loading conditions, curated from experimental investigations and academic literature [6-9, 21]. To ensure consistency and enhance the performance of machine learning models, the dataset was standardized, bringing all features to a comparable scale. Feature selection was guided by engineering relevance and statistical analysis, emphasizing parameters with the most significant impact on structural stability and axial load capacity. The dataset utilized in the development of the machine learning models is presented in Table 1.

**Table 1.** Description of the Dataset Used for Machine Learning Model Development

| No | References | Yield stress (Mpa) | Number of experiments | Testing method |
|---|---|---|---|---|
| 1 | Pala and Şenaysoy [22] | 261 | 40 | Buckling test |
| 2 | Kadi [23] | 355 | 7 | Buckling test |
| 3 | Crisan et. al. [24] | 463 | 10 | Buckling test |
| 4 | Bonada et. al. [25] | 418 | 1 | Buckling test |
| 5 | Özkal and Yazici [21] | 433 | 15 | Buckling test |
| 6 | Zhang and Alam [26] | 250 | 18 | Buckling test |
| 7 | Miyazaki et. al. [27] | 345 | 36 | Buckling test |
| 8 | Kumar and Jayachandran [28] | 365 | 3 | Buckling test |
| 9 | Cheng et. al. [7] | 442 | 8 | Buckling test |
| 10 | Zhang and Alam [26] | 470 | 3 | Buckling test |
| 11 | Liu et. al. [29] | 515 | 12 | Buckling test |
| 12 | Ren et. al. [11] | 390 | 54 | Buckling test |
| 13 | Koen [30] | 550 | 10 | Buckling test |
| 14 | Elias et. al. [6] | 369 | 44 | Buckling test |

## 2.2. Data Analysis and Visualization

To gain insights into the dataset, the distribution of structural parameters was first analyzed. This step allowed the identification of potential deviations from normality and provided an overview of the general statistical tendencies of the variables. To explore the relationships among features, a correlation matrix was computed using Pearson correlation coefficients, which quantify the strength and direction of linear relationships between pairs of variables. Pearson coefficients range from -1 (perfect negative linear correlation) to +1 (perfect positive linear correlation), with 0 indicating no linear correlation. This analysis helped identify which features are most strongly associated with the load-bearing capacity (P), enabling the selection of relevant predictors. To improve readability and reduce redundancy, the upper triangle of the matrix was masked. In addition, scatter plots were generated to visually examine the relationships between structural parameters and the target variable. These visualizations facilitated the detection of non-linear trends and potential outliers, offering deeper insights into the underlying data structure.

## 2.3. Model Development and Training Process

The development of machine learning (ML) models for predicting the axial load capacity of steel storage rack columns followed a structured workflow, including feature scaling, model selection, hyperparameter tuning, cross-validation, and performance evaluation [31]. Prior to model training, PowerTransformer (standardized=True) was applied to normalize the features, ensuring numerical stability and mitigating the effect of skewed data distributions. The dataset, consisting of structural and material parameters from 261 storage rack samples, was partitioned into training (80%) and test (20%) sets to ensure that the models were evaluated on unseen data. This approach prevents overfitting by ensuring that models do not simply memorize the training data but instead generalize well to new inputs. To further validate model performance, K-Fold Cross-Validation with K=5 was applied during training. This method divides the training dataset into five equal folds, iteratively using four folds for training and one fold for validation. The process is repeated five times, with each fold serving as the validation set once, ensuring that all data points contribute to both training and validation phases. The average of the performance metrics across folds was calculated to obtain a robust estimate of each model's generalization performance [9].

An extensive machine learning framework was implemented to predict the axial load-bearing capacity of storage rack columns using a broad set of regression algorithms. The models evaluated include ensemble-based methods as displayed in Table 1 [31]:

Table 1. Advantages and limitations of the used models

| Model | Advantages | Limitations |
| --- | --- | --- |
| **Gradient Boosting (GBR)** | High accuracy; robust to overfitting; handles complex non-linear relationships. | Computationally intensive; sensitive to hyperparameters; slow training on large datasets. |
| **Extra Trees (ETR)** | Reduces variance through randomization; faster than other ensemble methods. | Less interpretable; can overfit noisy data. |
| **Bagging Regressor** | Reduces variance; effective with high-variance models like decision trees. | Less effective on small datasets; limited improvement on low-bias models. |
| **Random Forest (RF)** | Robust; handles non-linearities; reduces overfitting; performs well with minimal tuning. | Limited interpretability; some redundancy among trees. |
| **XGBoost (XGB)** | Efficient, scalable; includes regularization and built-in handling of missing data. | Complex to tune; prone to overfitting if not properly regularized. |
| **AdaBoost** | Emphasizes hard-to-predict samples; relatively simple and interpretable. | Sensitive to noise and outliers; less effective with highly complex relationships. |
| **Linear Regression** | Simple; interpretable; fast to train; strong baseline. | Assumes linear relationships; poor performance with non-linearity or multicollinearity. |
| **Ridge Regression** | L2 regularization controls overfitting; handles multicollinearity well. | Does not perform feature selection; retains all predictors. |
| **Lasso Regression** | L1 regularization promotes sparsity; performs automatic feature selection. | Can be unstable with correlated predictors; may eliminate relevant features. |
| **ElasticNet** | Combines L1 and L2 regularization; balances sparsity and shrinkage; good for correlated features. | More complex tuning due to two regularization parameters. |
| **Bayesian Ridge** | Incorporates uncertainty; robust with multicollinearity; estimates posterior distributions. | Computationally more intensive than basic linear models. |
| **Support Vector Regressor (SVR)** | Effective for non-linear data with kernel trick; good generalization in high-dimensional spaces. | Sensitive to kernel choice and scaling; not scalable for large datasets. |
| **K-Nearest Neighbors (KNN)** | Simple; non-parametric; no training phase required. | Computationally heavy during prediction; sensitive to outliers and feature scaling. |
| **Decision Tree (DTR)** | Easy to interpret; handles both categorical and numerical data; fast to train. | Easily overfits; unstable with small data changes. |
| **Partial Least Squares (PLS)** | Good with small samples and multicollinearity; reduces dimensionality. | Less interpretable; may underperform on large, complex datasets. |

Each model was embedded within a ML pipeline that ensured consistent preprocessing (e.g., standardization) and efficient hyperparameter optimization using GridSearchCV [31, 32]. This approach ensured fair performance comparison and optimal tuning across all models. The main hyperparameters for each model are shown in table 2.

Table 2. The main hyperparameters tuned for each algorithm:

| Model | Key Tuned Hyperparameters |
|---|---|
| **Gradient Boosting** | n_estimators, learning_rate, max_depth, min_samples_split, min_samples_leaf |
| **Extra Trees** | n_estimators, max_depth, min_samples_split, min_samples_leaf, max_features |
| **Bagging Regressor** | n_estimators, max_samples, max_features, bootstrap, bootstrap_features |
| **Random Forest** | n_estimators, max_depth, min_samples_split, min_samples_leaf, max_features |
| **XGBoost** | n_estimators, learning_rate, max_depth, subsample, colsample_bytree, reg_alpha, reg_lambda |
| **AdaBoost** | n_estimators, learning_rate, loss |
| **Lasso** | alpha, max_iter, selection |
| **ElasticNet** | alpha, l1_ratio, max_iter |
| **Ridge** | alpha, solver |
| **Linear Regression** | (No hyperparameters – baseline model) |
| **Bayesian Ridge** | alpha_1, alpha_2, lambda_1, lambda_2 |
| **Support Vector Regressor (SVR)** | kernel, C, epsilon, gamma |
| **K-Nearest Neighbors** | n_neighbors, weights, algorithm |
| **Decision Tree** | max_depth, min_samples_split, min_samples_leaf, max_features |
| **Partial Least Squares (PLS)** | n_components |

---

The models were evaluated using standard regression metrics, including the coefficient of determination (R²) to measure the variance explained by the model, mean squared error (MSE) for quantifying average squared prediction errors, mean absolute error (MAE) to assess the average magnitude of prediction errors, and root mean squared error (RMSE) for capturing prediction accuracy more sensitively. Thus, coefficient of determination (R²) measures the proportion of variance in the target variable explained by the model. It provides an indication of how well the predicted values align with the actual values. An R² value close to 1 indicates a strong predictive model, while a value close to 0 suggests poor performance.

$$R^2 = 1 - \frac{\sum(y_i - \hat{y}_i)^2}{\sum(y_i - \bar{y})^2} \quad (1)$$

where $y_i$ is the actual value, $\hat{y}_i$ is the predicted value, and $\bar{y}$ is the mean of the actual values.

Mean Absolute Error (MAE): MAE represents the average magnitude of errors between the predicted and actual values. It is a straightforward measure that quantifies the absolute difference without penalizing larger errors disproportionately. Lower MAE values indicate higher accuracy.

$$MAE = \frac{1}{n}\sum_{i=1}^{n}|y_i - \hat{y}_i| \quad (2)$$

Root Mean Squared Error (RMSE): RMSE measures the square root of the mean of the squared errors. Unlike MAE, RMSE penalizes larger errors more heavily, making it particularly useful for identifying models with significant outliers. Smaller RMSE values signify better predictive performance.

$$RMSE = \sqrt{\frac{1}{n}\sum_{i=1}^{n}(y_i - \hat{y}_i)^2} \quad (3)$$

These metrics were selected to provide a comprehensive evaluation of model accuracy and error magnitude, ensuring that both overall performance and the impact of extreme deviations were considered.

The cross-validation results were recorded for each model, enabling a direct comparison of generalization performance. The cross-validation results confirmed that the ensemble model outperformed individual models in terms of prediction reliability and generalization. After evaluating all models, the best-performing model, achieving the highest validation accuracy. The trained model was saved using joblib for future applications [15], along with the selected feature set. The final model selection and training process followed a systematic approach, ensuring accurate, reliable, and generalizable predictions for the axial load capacity of steel storage rack columns.

**2.4. Buckling Tests Procedure: Validation dataset**

The buckling tests conducted on storage rack uprights aim to evaluate their axial load-bearing capacity and determine critical buckling modes under compressive loading. These tests typically involve cold-formed steel uprights with different cross-sectional geometries, lengths, and material properties. Both bolted and unbolted configurations are examined to assess the structural impact of mechanical fasteners. The experimental setup includes a universal testing machine equipped with hydraulic actuators applying axial compressive loads, as shown in Fig. 1. End conditions such as pinned or fixed supports are configured to replicate real-world installation scenarios. Loading plates are securely attached at both ends to ensure proper load transmission and minimize premature local failures. Axial loads are applied incrementally, while displacement sensors and load cells record lateral deformations and applied forces. The onset of critical buckling is indicated by significant lateral deflection. Various buckling modes such as local, distortional, flexural, and torsional-flexural buckling are observed and classified. The maximum axial load sustained before buckling failure is recorded for further analysis. The experimental findings consistently highlight the effectiveness of bolted connections in enhancing buckling resistance, increasing axial load capacity, and suppressing distortional buckling modes. These results underline the importance of mechanical fasteners in improving the overall stability and durability of storage rack systems [8, 21].

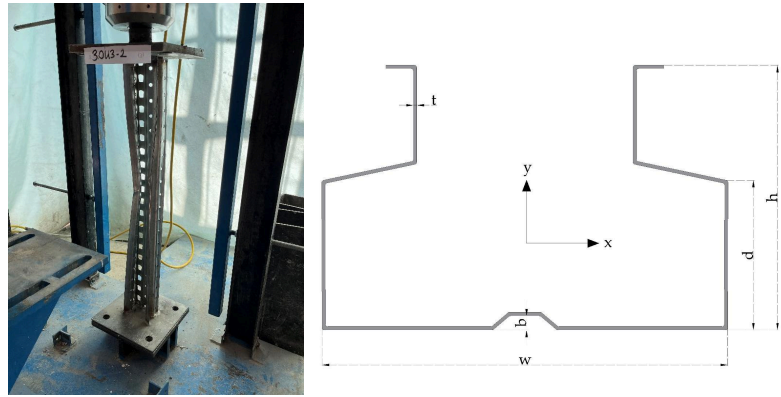

**Figure 1.** Buckling test setup and cross-sectional geometry of the steel storage rack column. The left panel shows the experimental setup used for the buckling tests. The right panel illustrates the cross-sectional geometry of the steel storage rack column, highlighting key structural parameters: w (back width), h (total height), b (indentation in the back section), and d (side fold distance). These geometric features play a critical role in determining the column's axial load capacity and are used as input variables in the ML models.

### 2.5. User Interface Development

To facilitate practical application of the predictive models, a graphical user interface (GUI) was developed using the Streamlit framework in Python by following the schematics of our approach displayed in Fig 2. This lightweight, interactive interface allows users to input design parameters either manually through form fields or by uploading a CSV file. The interface processes the input, applies the trained best model, and returns real-time predictions of axial load-bearing capacity. Visual feedback is provided via dynamically generated plots and tables, enhancing interpretability for both technical and non-technical users. The GUI is designed to be accessible, requiring no programming expertise, thereby bridging the gap between advanced ML models and day-to-day engineering decision-making. Predictions were displayed with two decimal precision in kilonewtons (KN). Error management features were also incorporated to ensure smooth system operation [10].

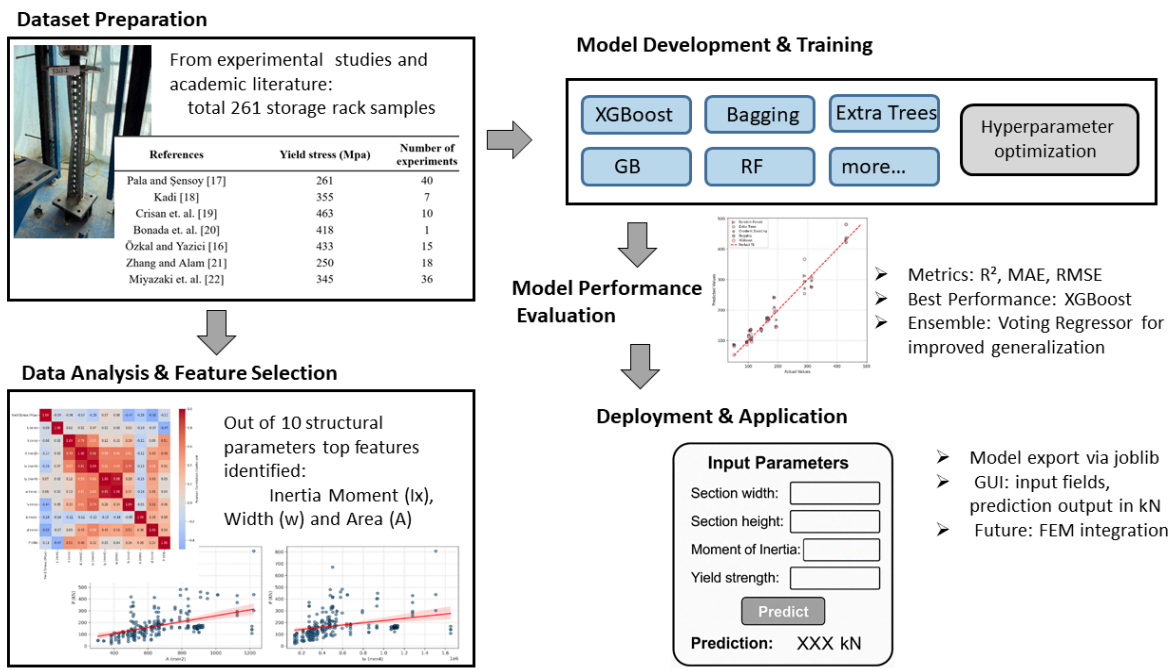

**Figure 2** Schematic representation of the methodology used to select the optimal stacked ML model and evaluate its performance

## 3. RESULTS

This section presents the results obtained from the ML models employed in the study to predict the critical buckling load of storage rack systems. The discussion is structured to address key findings from data distribution, correlation analysis, model comparison, and the prediction performance of the best-performing model. Each graphical representation provides insights into the relationships between structural parameters and their influence on buckling load predictions. Feature importance analysis was conducted to identify the most influential parameters affecting the buckling load prediction.

### 3.1. Data Distribution Analysis

To understand the variability and distribution of the input parameters, histograms with Kernel Density Estimation (KDE) overlays were generated (Figure 3). The distributions indicate that most features exhibit non-normal behavior, with significant skewness observed in parameters such as cross-sectional area (A), moment of inertia ($I_x$ and $I_y$), and yielding stress. These irregular distributions emphasize the need for robust models that can handle data variability without being biased by extreme values or outliers. The cross-sectional area (A) and moment of inertia ($I_x$) parameters demonstrate a bi-modal distribution, which suggests the presence of multiple structural categories within the dataset. The yielding stress parameter shows a concentrated distribution between 250 MPa and 300 MPa, consistent with standard structural steel properties used in practical applications. The non-uniform distribution of input variables confirms the necessity of using techniques such as feature scaling and transformations during the modeling process to improve predictive performance. To ensure data integrity, missing values were handled using appropriate imputation techniques. Additionally, outliers were detected using the Interquartile Range (IQR) method and removed to improve data consistency. Standardization was applied using the PowerTransformer technique to normalize the dataset and enhance model performance.

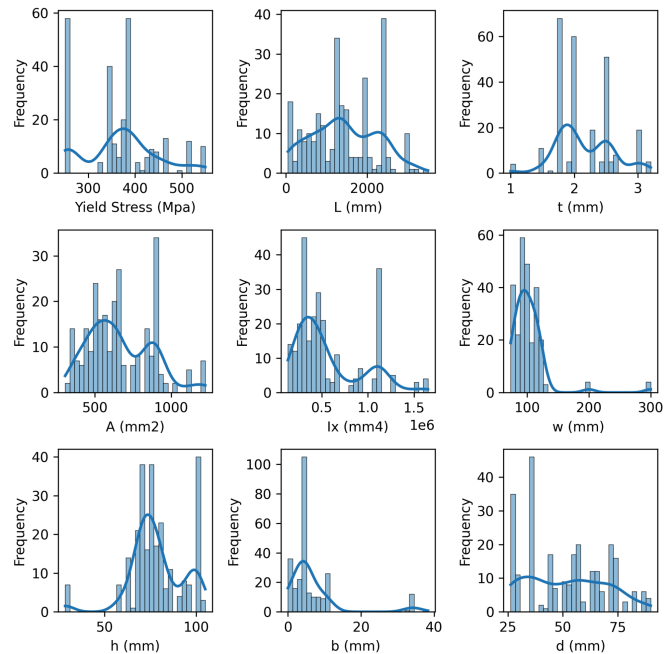

**Figure 3.** Feature distributions of the structural parameters with KDE overlays.

### 3.2. Correlation Analysis

The correlation matrix shown in Figure 4 highlights the relationships between structural parameters and the target variable, the critical buckling load $P/P_{cr}$. This matrix provides critical insights into parameter dependencies and their influence on the response variable. Moment of inertia ($I_x$) and width (w) exhibit the strongest positive correlation with $P/P_{cr}$ (0.91 and 0.95, respectively), emphasizing their dominant role in determining the load-bearing capacity. Conversely, parameters such as yielding stress and height (h) show weaker correlations with the critical buckling load. This suggests that while material properties are important, geometric factors like cross-sectional dimensions and inertia are the primary drivers of structural stability. Notably, there is little multicollinearity among input features, as most pairwise correlations fall below 0.8. This ensures that models are not adversely affected by redundant input information. The correlation results underline the importance of carefully selecting influential features when building predictive models, as demonstrated by the subsequent model performance analysis.

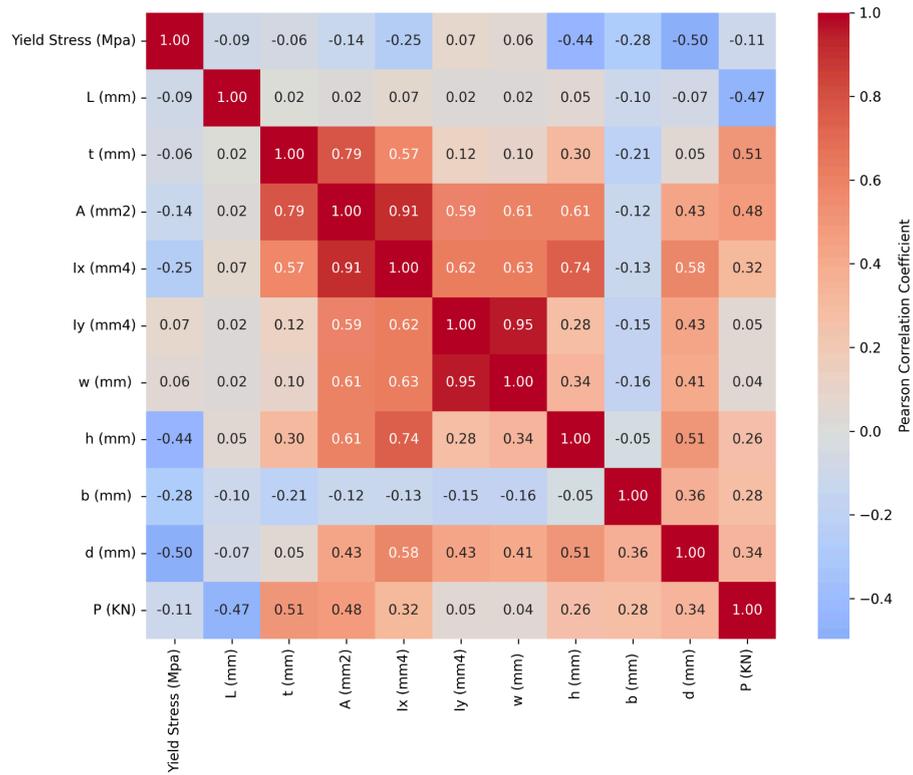

**Figure 4.** Correlation matrix of structural parameters and buckling load *P*.

### 3.3. Scatter Plot Analysis

Scatter plots between individual features and the target variable P/P$_{cr}$ (Figure 5) reveal additional relationships and potential trends. These plots allow a visual assessment of linearity or non-linearity between predictors and the response.

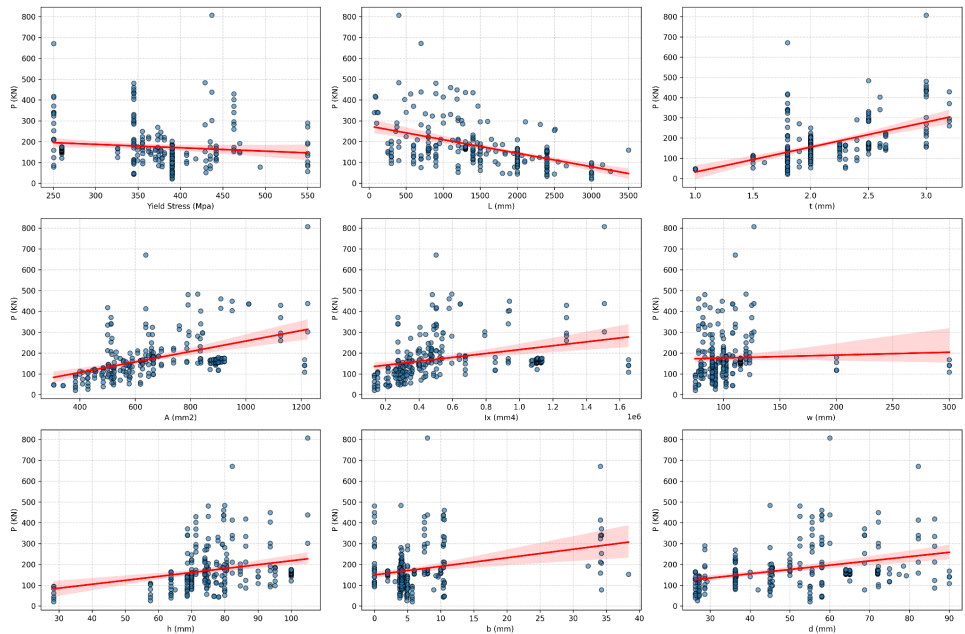

**Figure 5.** Scatter plots showing the relationships between individual features and the critical buckling load *P*.

The moment of inertia ($I_x$) and width (w) parameters show clear positive trends, confirming their strong correlation with $P/P_{cr}$. Higher inertia values are consistently associated with increased buckling loads. Features such as height (h) and thickness (t) exhibit more scattered relationships, indicating weaker and less linear associations with the critical load. Outliers are evident in parameters like $I_y$ and A, particularly in low $P/P_{cr}$ ranges, which could potentially affect model training. These findings confirm that while certain parameters are primary contributors, robust predictive models must account for both linear and non-linear relationships in the dataset.

### 3.4. Model Comparison

The comparative performance of the machine learning models is summarized in Figure 6. The comparative performance of the machine learning models is summarized in Figure 6, evaluated using R², Mean Squared Error (MSE), Mean Absolute Error (MAE), and Root Mean Squared Error (RMSE). Among the tested models, Gradient Boosting Regressor (GBR) achieved the highest predictive accuracy, with an R² score of 0.97 on the validation set, closely followed by Extra Trees Regressor (ETR) and Bagging Regressor. These results underscore the superior capability of ensemble-based learning approaches in modeling the non-linear and multivariate nature of structural load-bearing behavior.

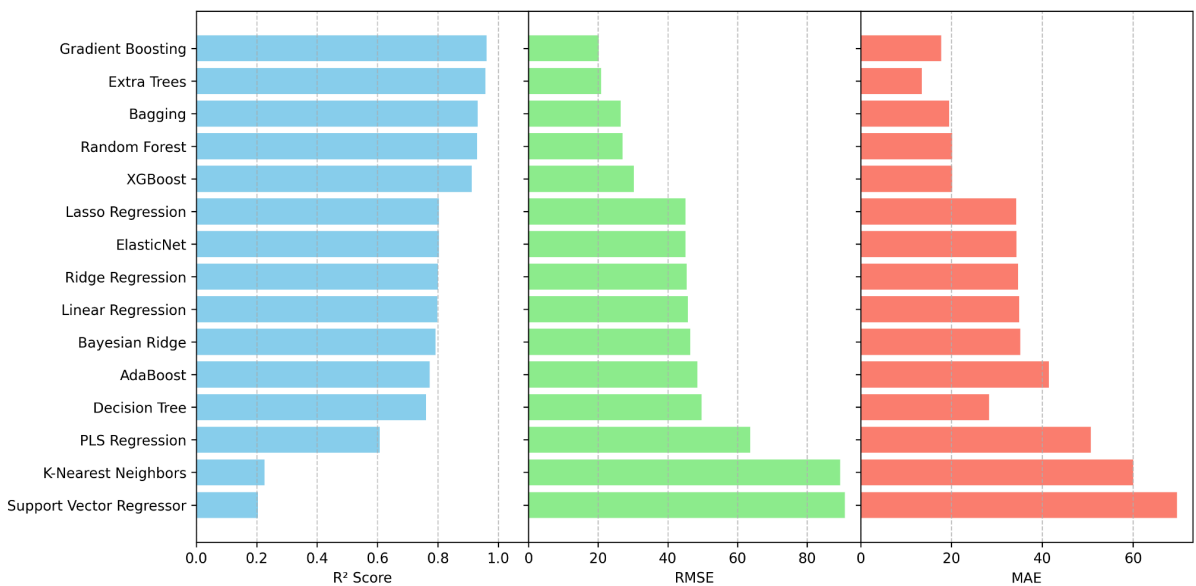

**Figure 6.** Model performance comparison.

Gradient Boosting outperformed all other models primarily due to its ability to sequentially reduce prediction errors by focusing on difficult-to-predict samples during training. Its built-in regularization mechanisms and shallow tree structures mitigate overfitting, offering strong generalization capabilities. Although GBR has a relatively slower training process due to its iterative nature, this is offset by its consistent accuracy across all evaluated metrics. Additionally, the lowest RMSE and MAE values observed for GBR confirm its ability to minimize both large and small prediction errors, making it the most reliable model overall. Extra Trees also performed exceptionally well, demonstrating high accuracy and robustness. Its success is attributed to its use of randomized thresholds for node splitting, which reduces variance and enhances generalization—particularly effective when working with high-dimensional and complex datasets. Compared to traditional decision tree models, ETR benefits from faster computation and improved resilience to noise, though it lacks the iterative correction process inherent to boosting methods like GBR.

In contrast, K-Nearest Neighbors (KNN) and Support Vector Regressor (SVR) exhibited the weakest performance among the tested models. KNN, while simple and non-parametric, suffers from several limitations in this context: it is highly sensitive to feature scaling, struggles with high-dimensional data (the "curse of dimensionality"), and has poor extrapolation ability, leading to degraded performance on unseen data. Similarly, SVR—despite its strong theoretical foundations—proved less effective due to its sensitivity to kernel choice and parameter tuning. Moreover, its scalability limitations and high computational cost further hindered its applicability to the current dataset, which includes diverse structural parameters with potential non-linear interactions.

To enhance interpretability and understand the contribution of each feature to the Gradient Boosting model's predictions, a SHAP (SHapley Additive exPlanations) summary plot was generated and displayed in Fig. 7. This plot provides both global and local interpretability by visualizing the magnitude and direction of each feature's impact on the model's output.

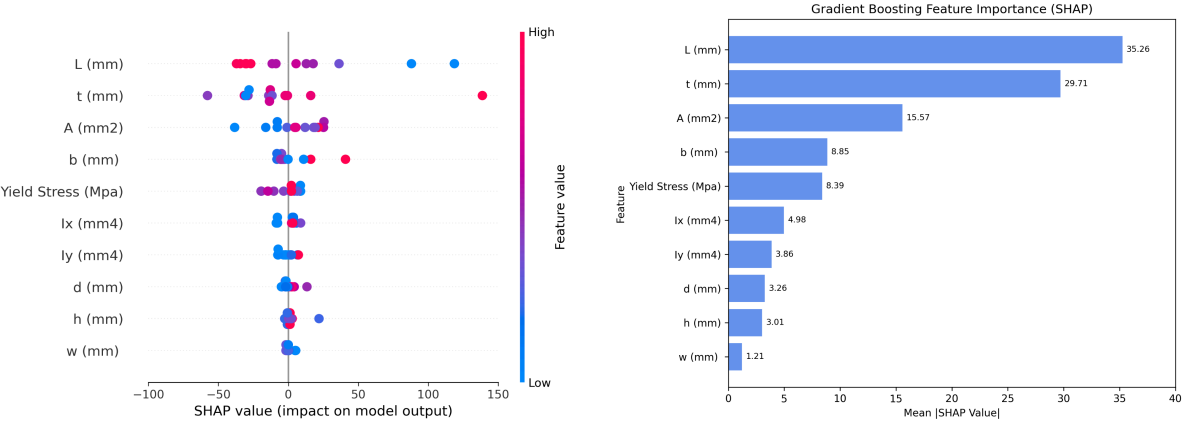

**Figure 7.** SHAP summary and importance plot considering Gradient boosting.

As shown in the SHAP summary and importance plot, the features L (length), t (thickness), A (cross-sectional area), and b (width) emerged as the most influential predictors of the load-bearing capacity. These features consistently appeared at the top of the plot, indicating their dominant contribution across the entire dataset. The mean absolute SHAP values, which quantify the average impact of each feature on the model output, reinforce this observation. Specifically: L (length) exhibited the highest mean SHAP value of 35.26, indicating that variations in this parameter most significantly affect the model's predictions; t (thickness) followed with a mean SHAP value of 29.71, suggesting its strong influence on structural stability; A (cross-sectional area) and b (width) had mean SHAP values of 15.57 and 8.85, respectively, also playing meaningful roles in the model's decision process. These results are consistent with engineering intuition, as geometric dimensions directly govern the axial stiffness and buckling behavior of structural components. The SHAP analysis not only validates the physical relevance of the selected features but also ensures transparency in the model's internal logic, making it a valuable tool for engineering design validation and decision-making.

To visually assess the predictive accuracy and generalization capability of the top-performing ML models, in Figure 8 we present the Actual vs. Predicted Values plot that is generated for the best models: Gradient Boosting, Bagging, Extra Trees, XGBoost, and Random Forest.

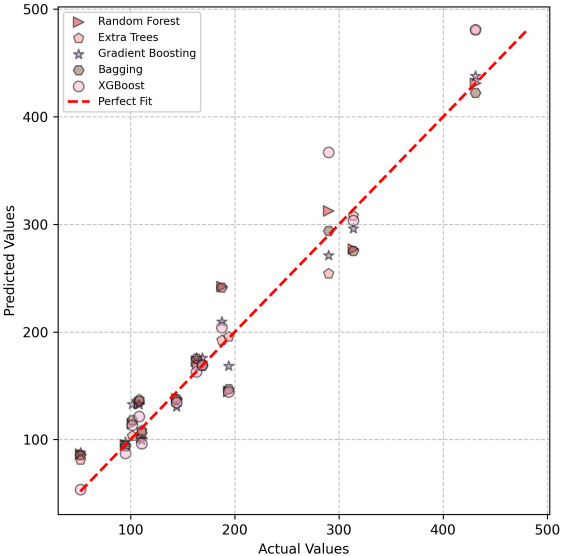

**Figure 8.** Actual vs. predicted values for the best models: Gradient Boosting, Bagging, Extra trees, XGBoost, and Random Forest.

The predicted load-bearing capacities are plotted against the true values from the validation dataset. A perfect prediction would align all data points along the diagonal reference line (y = x), indicating complete agreement between predicted and actual values. Among the models: Gradient Boosting showed the best alignment with the reference line, demonstrating excellent predictive performance and minimal dispersion. The data points are tightly clustered along the diagonal, reflecting its strong ability to capture complex relationships within the data; Extra Trees and XGBoost also performed well, with predicted values closely matching actual ones. Slight deviations from the diagonal line were observed, particularly at higher load values, suggesting a small degree of under- or overestimation; Random Forest delivered stable predictions with moderate variance. While its predictions follow the overall trend, there is slightly more scatter compared to Gradient Boosting and XGBoost, especially for mid-range values; Bagging exhibited the widest dispersion among the top models. Although it captures the general trend, greater variability is visible across the full range of predictions, likely due to its increased sensitivity to the training data distribution and weaker regularization. Overall, the Gradient Boosting Regressor consistently produced the most accurate and reliable predictions, as evidenced by the tight clustering around the reference line and the lowest RMSE and MAE values. These plots validate the model's suitability for predicting the axial load capacity of steel storage rack components and reinforce the effectiveness of ensemble-based boosting techniques for structural performance modeling.

### 3.5. Prediction Performance

To assess the generalizability of the Gradient Boosting Regressor—the best-performing model in this study—we validated its performance using an independent dataset composed of experimental measurements obtained in our laboratory. This validation step tested the model's ability to accurately predict the load-bearing capacity of storage rack components under real-world conditions. The comparison between predicted and actual values is illustrated in the Actual vs. Predicted Values plot in Figure 9, which shows a strong linear correlation, with data points closely aligned along the y = x reference line. The model achieved an excellent coefficient of determination ($R^2 = 0.96$), demonstrating its robustness and predictive accuracy. Furthermore, the root mean squared error (RMSE) and mean absolute error (MAE) values remained low, reinforcing the model's ability to generalize effectively beyond the training data. The majority of predictions are accurate, particularly in lower and mid-range buckling loads, reflecting the model's ability to generalize well across most of the dataset. A slight deviation is observed at higher load values, where the model underestimates the true axial load capacity. This discrepancy can be attributed to the limited number of high-capacity samples in the dataset, which restricts the model's learning in this range. To facilitate practical application of the developed machine learning model, a user-friendly graphical interface was implemented. This interface allows users to interact with the trained Gradient Boosting model and obtain axial load predictions for steel storage rack columns. As shown in Figure 9 (right panel), the interface provides two options: Single Prediction Mode: The user can manually input the structural parameters (e.g., length L, thickness t, cross-sectional area A, and width *b* and more features) into designated input fields. This mode is intended for quick evaluation of individual cases without the need for pre-formatted files; Batch Prediction Mode: Users can upload a .csv file containing multiple sets of structural features. This option enables batch processing and the prediction of load-bearing capacity for multiple samples simultaneously, making it particularly useful for evaluating experimental datasets or parametric studies. Once the input is provided, the interface automatically processes the data, applies the necessary transformations, and displays the predicted buckling loads in real time. This design ensures accessibility for engineers and practitioners, even those with no prior programming experience, streamlining the use of machine learning in practical structural design workflows.

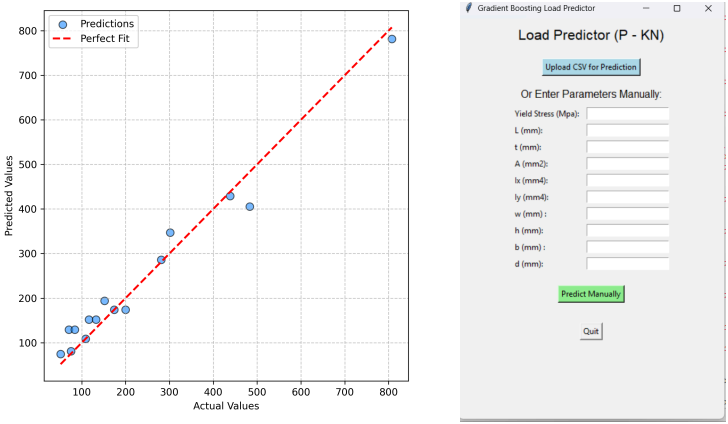

**Figure 9.** Validation of Gradient Boosting Model Using Experimental Data to predict Load-Bearing Capacity (left panel). graphical user interface (GUI) to compute the load from CSV file or a single point calculation.

A summary of the optimized hyperparameters used for this task is presented in Table 3, highlighting the importance of fine-tuning parameters such as learning rate, maximum tree depth, number of estimators, and minimum samples per leaf. These results confirm that the Gradient Boosting model can reliably support structural design predictions, bridging the gap between machine learning modeling and experimental validation.

Table 3. Hyperparameters of Gradient Boosting model

| Variations | Value | Optimal |
| --- | --- | --- |
| Number of boosting stages (n_estimators) | 100, 200, 300 | 200 |
| Learning rate | 0.01, 0.1, 0.3 | 0.1 |
| Maximum depth of the individual estimators | 3, 5, 7 | 5 |
| Minimum samples required to split a node | 2, 5, 10 | 2 |
| Minimum samples required at each leaf node | 1, 2, 5 | 1 |
| Subsample | 0.8, 0.9, | 1 |

These findings reinforce the applicability of advanced machine learning techniques for structural performance evaluation and predictive modeling. The insights derived from this study can be utilized to optimize the design of storage rack systems, improve load-bearing predictions, and enhance overall structural safety and efficiency. Thus, the integration of machine learning methods has enabled a detailed and accurate assessment of critical buckling loads. Future studies may focus on expanding the dataset, incorporating additional structural parameters, and exploring deep learning methods to further enhance prediction accuracy.

## 4. CONCLUSION AND FUTURE RESEARCH

This study presented a comprehensive investigation into the use of machine learning techniques for predicting the critical buckling load of steel storage rack systems. By employing a dataset of 261 samples characterized by ten geometric and material features, we demonstrated the strong potential of data-driven models to solve complex structural engineering problems—particularly in contexts where traditional analytical or empirical methods are limited by geometric complexity or variable loading conditions. Among the 15 models tested, ensemble-based boosting approaches—especially Gradient Boosting Regressor and Extra Trees Regressor—consistently delivered the best predictive performance, achieving the lowest RMSE and MAE values and an $R^2$ of 0.97 on the validation set. These models proved effective at handling high-dimensional, skewed, and nonlinear data, making them well-suited for engineering applications where interactions between features are complex and often non-intuitive. Model interpretability analyses revealed that geometric features, particularly length (L), thickness (t), cross-sectional area (A), and width (b), were the most influential predictors of buckling load. This aligns with classical mechanics principles, where these parameters significantly affect axial stiffness and stability. Material properties, such as yield strength, were found to have a secondary but measurable influence, underscoring the multifactorial nature of buckling resistance. Supporting visualizations, including scatter plots and correlation matrices, enhanced understanding of the feature-target relationships and helped identify potential outliers and data inconsistencies. Furthermore, the model's performance was validated using experimental data, where the Gradient Boosting model achieved an $R^2$ of 0.96, confirming its ability to generalize to real-world scenarios. Despite these promising results, slight underestimations were observed at higher load values, indicating the need to extend the dataset to include more high-load cases.

While the current study presents promising results, several opportunities for further research and development remain:

- **Dataset Expansion and Diversity:** Expanding the dataset to include a larger number of samples, particularly for structures with higher buckling loads and different material types, will enhance the robustness and generalization of machine learning models. Incorporating experimental data from real-world tests across various load and boundary conditions can further validate and improve model accuracy.

- **Integration with Structural Health Monitoring (SHM):** Incorporating real-time sensor data into machine learning models can facilitate predictive maintenance and structural health monitoring of storage rack systems. Such systems can identify early signs of failure or degradation, enabling timely interventions and reducing the risk of catastrophic failures.
- **Energy and Cost Optimization:** Future studies can explore the integration of machine learning models with energy-efficient and cost-optimized storage system designs. For instance, algorithms can optimize material usage, structural dimensions, and manufacturing costs while maintaining stability and load-bearing requirements.

In conclusion, the findings of this study demonstrate the immense potential of machine learning for the prediction and optimization of storage rack systems' buckling performance. By combining advanced algorithms with robust structural engineering principles, this approach paves the way for smarter, more efficient, and safer storage solutions. The proposed future research directions aim to build upon this foundation, addressing existing limitations while further expanding the applicability of machine learning in structural engineering. Through continued development, these techniques can redefine the standards for structural assessment and design, offering significant benefits for both academia and industry.

**Declaration of Competing Interest**

The authors declare the following financial interests/personal relationships which may be considered as potential competing interests: One of our authors reports financial support was provided by the National Centre for Nuclear Research. If there are other authors, they declare that they have no known competing financial interests or personal relationships that could have appeared to influence the work reported in this paper.

**Acknowledgements**

We acknowledge support from the European Union Horizon 2020 research and innovation program under grant agreement no. 857470 and from the European Regional Development Fund via the Foundation for Polish Science International Research Agenda PLUS program grant No. MAB PLUS/ 2018/8.

**Data availability**

Data will be made available on request.